\documentclass[11pt,a4paper]{article}
\usepackage[hyperref]{acl2020}
\usepackage{times}
\usepackage{latexsym}

\usepackage{microtype}

\aclfinalcopy 

\usepackage{adjustbox}
\usepackage{url}
\usepackage{color}
\usepackage{multirow}
\usepackage{algorithm}

\usepackage{amsmath}

\usepackage{makecell}
\usepackage{graphicx}
\usepackage{caption}
\usepackage{subcaption}
\usepackage{amssymb}

\usepackage{arydshln}

\newcommand{\LeakyReLU}{{\rm{LeakyReLU}}}

\newcommand{\shprivate}{{\rm{shprivate}}}
\newcommand{\dynfusion}{{\rm{dynamic}}}
\title{
	Dynamic Fusion Network for Multi-Domain End-to-end Task-Oriented Dialog}
\author{Libo Qin$^{\dag}$, Xiao Xu$^{\dag}$, Wanxiang Che$^{\dag}$\thanks{\ \ Email corresponding.}, Yue Zhang$^{\ddag}$, Ting Liu$^{\dag}$ \\
	$^\dag$Research Center for Social Computing and Information Retrieval \\
	$^\dag$Harbin Institute of Technology, China \\
	$^\ddag$School of Engineering, Westlake University, China \\
		$^\ddag$Institute of Advanced Technology, Westlake Institute for Advanced Study \\
	 $^\dag$\{lbqin, xxu, car,tliu\}@ir.hit.edu.cn,
		yue.zhang@wias.org.cn\
}

\date{}

\begin{document}
\maketitle
\begin{abstract}
Recent studies have shown remarkable success in end-to-end task-oriented dialog system. 
However, most neural models rely on large training data, which are only available for a certain number of task domains, such as navigation and scheduling.
 This makes it difficult to scalable for a new domain with limited labeled data. 
However, there has been relatively little research on how to effectively use data from all domains to improve the performance of each domain and also unseen domains. 
To this end, we investigate methods that can make explicit use of domain knowledge and introduce a shared-private network to learn shared and specific knowledge. In addition, we propose a novel Dynamic Fusion Network (DF-Net) which automatically exploit the relevance between the target domain and each domain.
Results show that our model outperforms existing methods on multi-domain dialogue, giving the state-of-the-art in the literature. 
Besides, with little training data, we show its transferability by outperforming prior best model by 13.9\% on average.
\end{abstract}
\section{Introduction} \label{sec:intro}
\begin{figure}[t]
	\centering
	\includegraphics[scale=0.32]{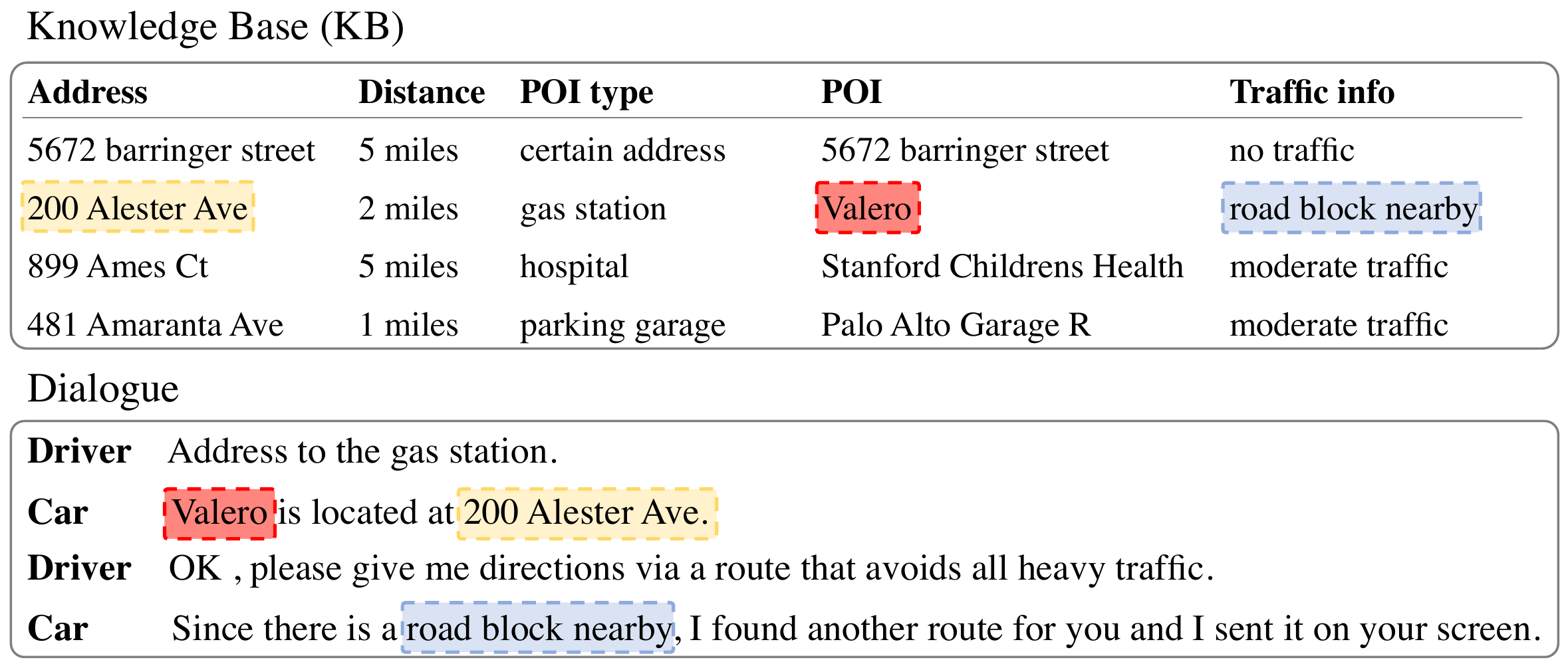}
	\caption{
		Example of a task-oriented dialogue that incorporates a knowledge base (KB) from the SMD dataset \cite{eric-etal-2017-key}. 
		Words with the same color refers queried entity from the KB.
		Better viewed in color.
	}
	\label{fig:example}
\end{figure}

\textit{Task-oriented dialogue systems} \cite{DBLP:journals/pieee/YoungGTW13} help users to achieve specific goals such as restaurant reservation or navigation inquiry.
In recent years, 
end-to-end methods
in the literature usually take the sequence-to-sequence (Seq2Seq) model to generate a response from a dialogue history \cite{eric-manning-2017-copy,eric-etal-2017-key,madotto-etal-2018-mem2seq,wen-etal-2018-sequence,gangi-reddy-etal-2019-multi,qin-etal-2019-entity,DBLP:conf/iclr/WuSX19}.
Taking the dialogue in Figure~\ref{fig:example} as an example, to answer the driver’s query about the ``\texttt{gas station}'', the end-to-end dialogue system directly generates system response given the query and a corresponding knowledge base (KB). 

Though achieving promising performance, end-to-end models rely on a considerable amount of labeled data, which limits their usefulness for new and extended domains.
In practice, we cannot collect rich datasets for each new domain.
Hence, it is important to consider
methods that can effectively  transfer knowledge from a source domain with sufficient labeled data to a target domain with limited or little labeled data.
\begin{figure*}[t]
	\centering
	\includegraphics[scale=0.4]{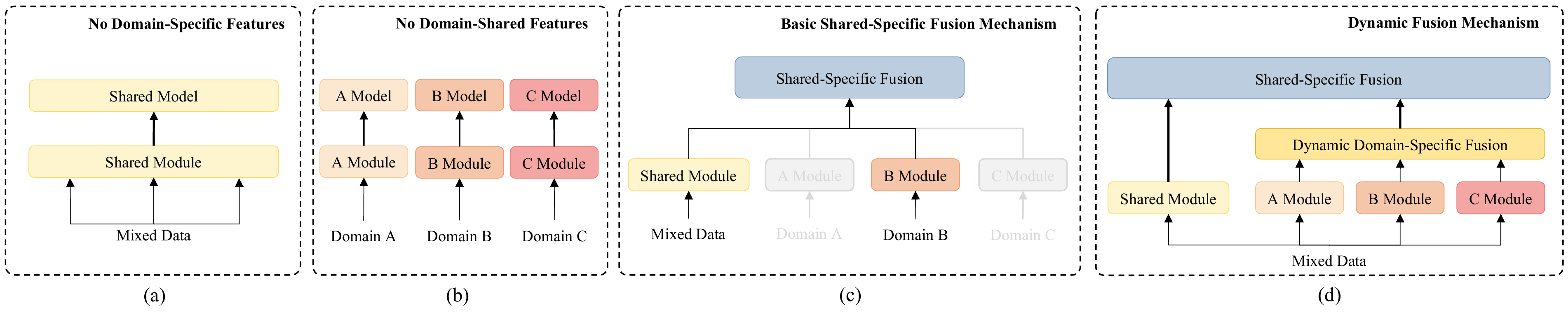}
	\caption{
		Methods for multi-domain dialogue.
		Previous work either trains a general  model on mixed multi-domain mixed datasets (a), or on each domain separately (b).
		The basic shared-private framework is shown (c).
		Our proposed extension with dynamic fusion mechanism is shown (d).
	}
	\label{fig:contract}
\end{figure*}

Existing work can be classified into two main categories.
As shown in Figure~\ref{fig:contract}(a), the first strand of work \cite{eric-manning-2017-copy,eric-etal-2017-key,madotto-etal-2018-mem2seq,DBLP:conf/iclr/WuSX19} simply combines multi-domain datasets for training.
Such methods can implicitly extract the shared features but fail to effectively capture domain-specific knowledge.
As shown in Figure~\ref{fig:contract}(b),
The second strand of work \cite{wen-etal-2018-sequence,qin-etal-2019-entity} trains model separately for each domain, which can better capture domain-specific features.
However, those methods ignore shared knowledge between different domains (e.g. the \texttt{location} word exists in both schedule domain and navigation domain).

We consider addressing the limitation of existing work by modeling knowledge connections between domains explicitly.
In particular, a simple baseline to incorporate domain-shared and domain-private features is shared-private framework \cite{liu-etal-2017-adversarial,zhong-etal-2018-global, DBLP:conf/nlpcc/WuZJXW19}. Shown in Figure~\ref{fig:contract}(c), it includes a shared module to capture domain-shared feature and a private module for each domain.
The method explicitly differentiates shared and private knowledge.
However, this framework still has two issues: (1) given a new domain with extremely little data, the private module can fail to effectively extract the corresponding domain knowledge.
(2) the framework neglects the fine-grained relevance across certain subsets of domains. (e.g. schedule domain is more relevant to the navigation than to the weather domain.)

To address the above issues, we further propose a novel Dynamic Fusion Network (DF-Net), which is shown in Figure~\ref{fig:contract} (d). 
In contrast to the shared-private model, a dynamic fusion module (see \S\ref{sec:moe})  is further introduced to explicitly capture the correlation between domains.
In particular, a gate is leveraged to automatically find the correlation between a current input and all domain-specific models, so that a weight can be assigned to each domain for extracting knowledge. Such a mechanism is adopted for both the encoder and the decoder, and also a memory module to query knowledge base features. Given a new domain with little or no training data, our model can still make the best use of existing domains, which cannot be achieved by the baseline model.

We conduct experiments on two public benchmarks, namely SMD \cite{eric-etal-2017-key} and MultiWOZ 2.1 \cite{budzianowski-etal-2018-multiwoz}.
Results show that our framework consistently and significantly outperforms the current state-of-the-art methods.
With limited training data, our framework outperforms the prior best methods by 13.9\% on average. 

To our best of knowledge, this is the first work to  effectively explore shared-private framework in multi-domain end-to-end task-oriented dialog.
In addition, 
when given a new domain which with few or zero shot data, our extended dynamic fusion framework can utilize fine-grained knowledge to obtain desirable accuracies, which makes it more adaptable to new domains.

All datasets and code are publicly available at: \url{https://github.com/LooperXX/DF-Net}.

\section{Model Architecture} \label{basic}
\begin{figure}[t]
	\centering
	\includegraphics[scale=0.65]{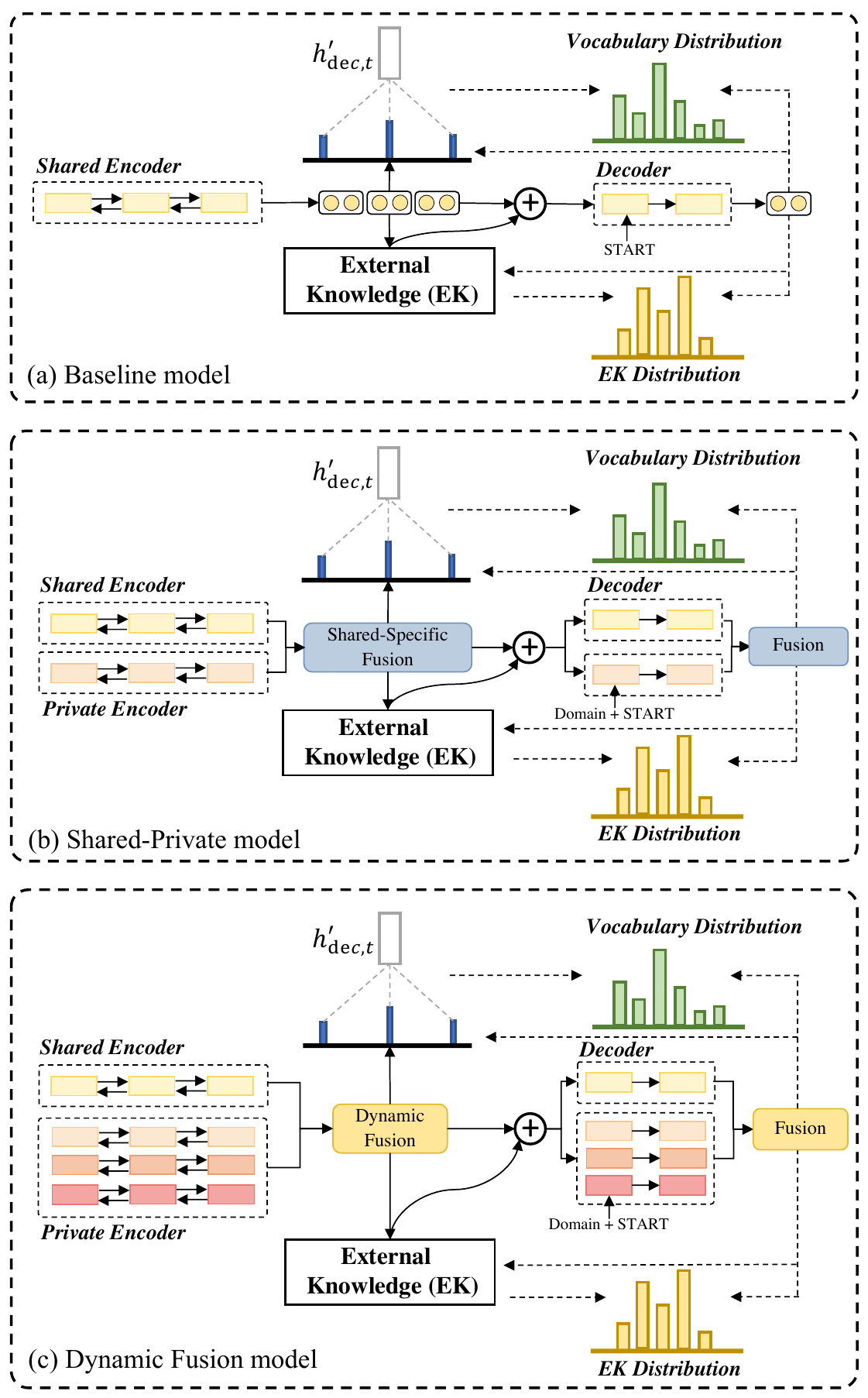}
	\caption{
		Workflow of our baseline and our proposed model.
	}
	\label{fig:framework}
\end{figure}

We build our model based on a seq2seq dialogue generation model (\S\ref{sec:seq2seq}), as shown in Figure \ref{fig:framework}(a).
To explicitly integrate domain awareness,
as shown in Figure \ref{fig:framework}(b) we first propose to use a shared-private framework (\S\ref{sec:shared-private}) to learn shared and the corresponding domain-specific features.
Next, we further use a dynamic fusion network (\S\ref{sec:moe}) to dynamically exploit the correlation between all domains for fine-grained knowledge transfer, which is shown in Figure \ref{fig:framework}(c).
In addition, adversarial training is applied to encourage shared module generate domain-shared feature.

\subsection{Seq2Seq Dialogue Generation} \label{sec:seq2seq}
We define the Seq2Seq task-oriented dialogue generation
as finding the system response ${Y}$
according to the input dialogue history $X$ and KB $B$.
Formally, the probability of a response is defined as
\begin{eqnarray}
p({Y} \mid X, B) = \prod_{t=1}^{n}  p (y_t \mid  y_1, ..., y_{t - 1}, X, B),
\end{eqnarray}
where $y_t$ represents an output token.
In a vanilla Seq2Seq task-oriented dialogue system \cite{eric-manning-2017-copy}, a long short-term Memory network (LSTM, \newcite{hochreiter1997long})  is used to encode the dialogue history
$X$ = (${x}_{1}, {x}_{2},.., {x}_{T}$)  ($T$ is the 
number of tokens in the dialogue history) to produce shared context-sensitive  hidden states $\boldsymbol{H}$ =  $(\boldsymbol{h}_{1}, \boldsymbol{h}_2, ..., \boldsymbol{h}_{T})$:
\begin{eqnarray}
	\boldsymbol{h}_{i}=\operatorname{BiLSTM}_\text{enc}\left(\phi^{emb}(x_{i}), \boldsymbol{h}_{i-1}\right),
\end{eqnarray}
where $\phi^{emb}(\cdot)$ represents the word embedding matrix. LSTM is also used to repeatedly predict outputs $(y_{1}, y_2, ... ,y_{t-1})$ by the decoder hidden states $({\boldsymbol{h}}_{\text{dec}, 1}, {\boldsymbol{h}}_{\text{dec}, 2}, ... ,{\boldsymbol{h}}_{\text{dec}, t})$.
For the generation of $y_t$, the model
first calculates an attentive representation ${\boldsymbol{h}}^{'}_{\text{dec}, t}$ of the dialogue history over the encoding representation $\boldsymbol{H}$.
Then, the concatenation of  ${\boldsymbol{h}}_{\text{dec}, t}$
and  ${\boldsymbol{h}}^{'}_{\text{dec}, t}$
is projected to the vocabulary space $\mathcal{V}$ by $\boldsymbol{U}$:
\begin{equation}
\label{vanilla}
	\boldsymbol{o}_t = \boldsymbol{U}\ [{\boldsymbol{h}}^{}_{\text{dec}, t}, {\boldsymbol{h}}^{'}_{\text{dec}, t}], 
\end{equation}
where $\boldsymbol{o}_t$ is the score (logit) for the next token generation.
The probability of next token $y_t \in \mathcal{V}$ is finally calculated as:
\begin{equation} \label{generation}
	p(y_t \mid y_1, ..., y_{t-1}, {X}, B) = \text{Softmax}(\boldsymbol{o}_t).
\end{equation}
Different from typical text generation with Seq2seq model, the successful conversations for task-oriented dialogue system heavily depend on accurate knowledge base (KB) queries.
We adopt the global-to-local memory pointer mechanism (GLMP) \cite{DBLP:conf/iclr/WuSX19} to query the entities in KB, which has shown the best performance. 
An external knowledge memory is proposed to store  knowledge base (KB) $B$ and dialogue history $X$.
The KB memory is designed for the knowledge source while the dialogue memory is used for directly copying history words.
The entities in external knowledge memory are represented in a triple format and stored in the memory module, which can be denoted as $M = [B; X] = (m_1,\dots,m_{b+T})$, where $m_i$ is one of the triplet of $M$, $b$ and $T$ denotes the number of KB and dialog history respectively.
For a $k$-hop memory network, the external knowledge is composed of a set of trainable embedding matrices $\boldsymbol{C} = (\boldsymbol{C}^1,\dots,\boldsymbol{C}^{k+1})$.
We can query knowledge both in encoder and decoder process to enhance model interaction with knowledge module.

\paragraph{Query Knowledge in Encoder} \label{encoder_query}
We adopt the last hidden state as the initial query vector:
\begin{equation}
\label{initial}
\boldsymbol{q}_\text{enc}^{1} = \boldsymbol{h}_{T}.
\end{equation}
In addition, it can loop over $k$ hops and compute the attention weights at each hop $k$ using
\begin{equation}
\boldsymbol{p}^k_i = \operatorname{Softmax}((\boldsymbol{q}_\text{enc}^k)^\top  \boldsymbol{c}^k_i),
\label{attn_eq}
\end{equation}
where $\boldsymbol{c}^k_i $ is the embedding in $i^{th}$ memory position using the embedding matrix $\boldsymbol{C}^k$.
We obtain the global memory pointer $G = (g_1,\dots,g_{b+T})$ by applying $\boldsymbol{g}^k_i = \operatorname{Sigmoid}((\boldsymbol{q}_\text{enc}^k)^\top \boldsymbol{c}^k_i)$, which is used to filter the external knowledge for relevant information for decoding.

Finally, the model reads out the memory $\boldsymbol{o}^k$ by the weighted sum over $\boldsymbol{c}^{k+1}$ and updates the query vector $\boldsymbol{q}_\text{enc}^{k+1}$. Formally,
\begin{equation}
\boldsymbol{o}_\text{enc}^k = \sum_i \boldsymbol{p}^k_i \boldsymbol{c}^{k+1}_i, \quad \boldsymbol{q}_\text{enc}^{k+1} = \boldsymbol{q}_\text{enc}^{k} + \boldsymbol{o}_\text{enc}^{k}.
\end{equation}

$\boldsymbol{q}_\text{enc}^{k+1}$ can be seen as the encoded KB information, and is used to initialized the decoder.
\paragraph{Query Knowledge in Decoder} \label{decoder_query}
we use a sketch tag to denote all the possible slot types that start with a special token. (e.g., \textit{@address} stands for all the \texttt{Address}).
When a sketch tag is generated by Eq.~\ref{generation} at $t$ timestep, we use the concatenation of the hidden states ${\boldsymbol{h}}_{\text{dec}, t}$
and the attentive representation ${\boldsymbol{h}}^{'}_{\text{dec},t}$ to query knowledge, which is similar with the process of querying knowledge in the encoder:
\begin{eqnarray}
\label{decoder_initial}
 \boldsymbol{q}_\text{dec}^{1} &=& [\boldsymbol{h}^{}_{\text{dec}, t}, \boldsymbol{h}^{'}_{\text{dec}, t}], \\
p^k_i &=& \text{Softmax}((\boldsymbol{q}_\text{dec}^k)^\top \boldsymbol{c}^k_i g^k_i).
\end{eqnarray}

Here, we can treat $P_{t}$ = ($p^k_1$,\dots,$p^{k}_{b+T}$) as the probabilities of queried knowledge, and select the word with the highest probability from the query result as the generated word. 

\subsection{Shared-Private Encoder-Decoder Model} \label{sec:shared-private}
The model in section~\ref{sec:seq2seq} is trained over mixed multi-domain datasets and the model parameters are shared across all domains.
We call such model as \textit{shared encoder-decoder} model.
Here, we propose to use a shared-private framework including a shared encoder-decoder for capturing domain-shared feature and a \textit{private} model for each domain to consider the domain-specific features explicitly.
Each instance ${X}$ goes through both the shared and its corresponding private encoder-decoder.
\paragraph{Enhancing Encoder}
Given an instance along with its domain, the shared-private encoder-decoder generates a sequence of encoder vectors denoted as $\boldsymbol{H}_{\text{enc}}^{\left\{ s,d \right\}}$,  including \textbf{s}hared and \textbf{d}omain-specific representation from corresponding encoder:
\begin{eqnarray}
\label{Eq:private_encoder}
\begin{aligned}
\boldsymbol{H}_{\text{enc}}^{\left\{ s,d \right\}}&\!=\!(\boldsymbol{h}_{\text{enc}, 1}^{\left\{ s,d \right\}},\dots,\boldsymbol{h}_{\text{enc}, T}^{\left\{ s,d \right\}})\\ 
&\!=\!\operatorname{BiLSTM}_{\text{enc}}^{\left\{ s,d \right\}}({X}).
\end{aligned}
\end{eqnarray}
\begin{figure}[t]
	\centering
	\includegraphics[scale=0.25]{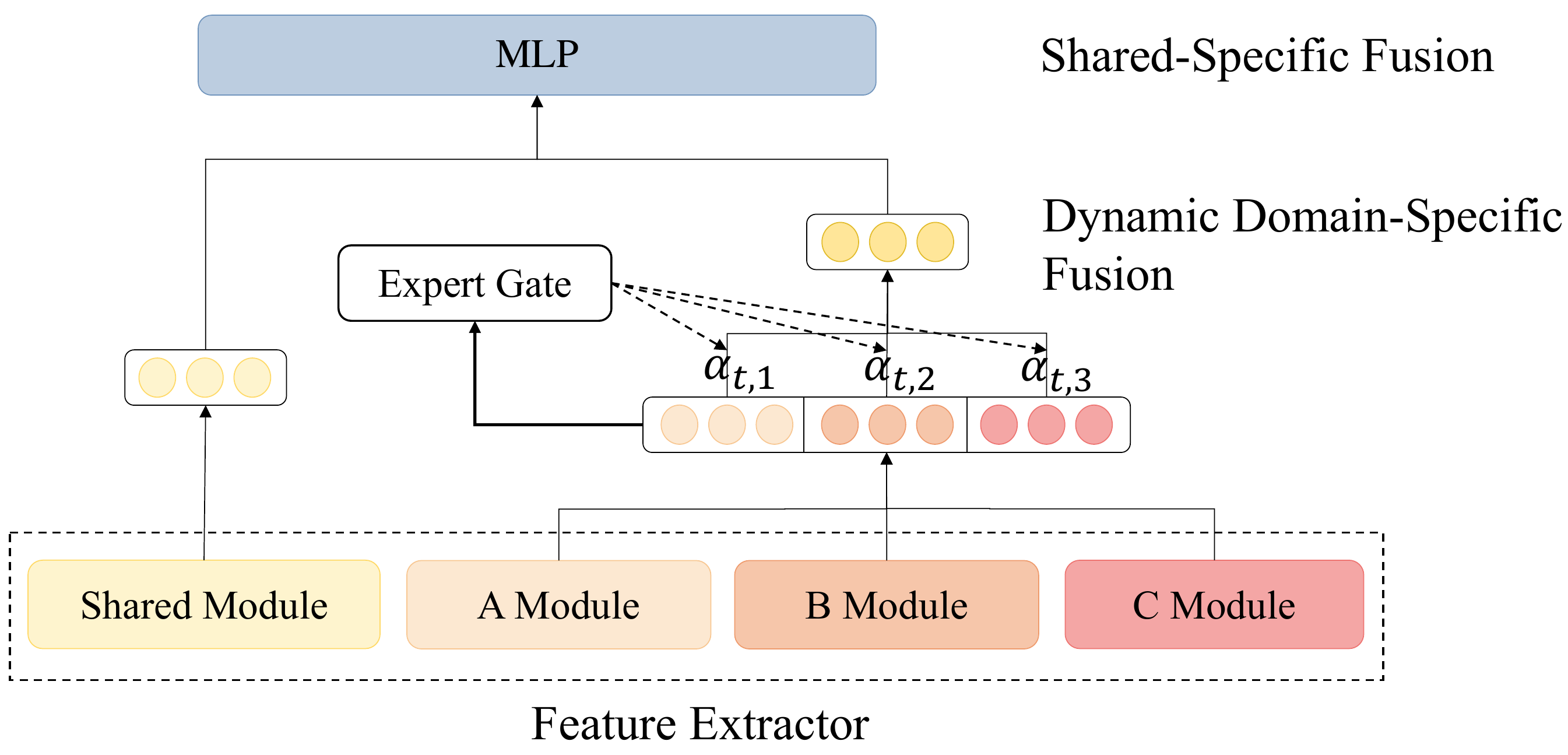}
	\caption{
		The dynamic fusion layer for fusing domain-shared feature and domain-specific feature.
	}
	\label{fig:knowledge_transfer}
\end{figure}
The final shared-specific encoding representation $	\boldsymbol{H}_{\text{enc}}^{f}$ is a mixture:
\begin{eqnarray}
	\boldsymbol{H}_{\text{enc}}^{f}\!=\!\boldsymbol{W}_{2}(\LeakyReLU(\boldsymbol{W}_{1}[\boldsymbol{H}_{\text{enc}}^{s}, \boldsymbol{H}_{\text{enc}}^{d}])).
\end{eqnarray}
For ease of exposition, we define the shared-specific fusion function as:
\begin{eqnarray}
\shprivate : (\boldsymbol{H}_{\text{enc}}^{s}, \boldsymbol{H}_{\text{enc}}^{d})  \rightarrow \boldsymbol{H}_\text{enc}^{f}.
\end{eqnarray}
In addition, self-attention has been shown useful for obtaining context information \cite{zhong-etal-2018-global}.
Finally, we follow \newcite{zhong-etal-2018-global} to use self-attention over $\boldsymbol{H}_{\text{enc}}^{f}$ to get context vector $\boldsymbol{c}_{\text{enc}}^{f}$.
We replace $\boldsymbol{h}_{T}$ with $\boldsymbol{c}_{\text{enc}}^{f}$ in Eq.~\ref{initial}.
This makes our query vector combine the domain-shared feature with domain-specific feature.

\paragraph{Enhancing Decoder}
At $t$ step of the decoder, the private and shared hidden state is:
\begin{eqnarray}
\label{Eq:private_encoder}
\boldsymbol{h}_{{\text{dec}, t}}^{\left\{ s,d \right\}}= \operatorname{LSTM}_{{\text{dec}, t}}^{\left\{ s,d \right\}}({X}).
\end{eqnarray}

We also apply the shared-specific fusion function to the hidden states and the mixture vector is:
\begin{eqnarray}
	\shprivate : (\boldsymbol{h}_{{\text{dec}, t}}^{s}, \boldsymbol{h}_{{\text{dec}, t}}^{d}) \rightarrow \boldsymbol{h}_{{\text{dec}, t}}^{f}.
\end{eqnarray}
Similarly, we obtain the fused attentive representation ${\boldsymbol{h}}^{f'}_\text{dec,t}$ by applying attention from  $\boldsymbol{h}_{\text{dec}, t}^{f}$ over $\boldsymbol{H}_\text{enc}^{f}$.
Finally, we replace $[\boldsymbol{h}^{}_{\text{dec}, t}, \boldsymbol{h}^{'}_{\text{dec}, t}]$ in Eq.~\ref{decoder_initial} with $[ \boldsymbol{h}_{\text{dec}, t}^{f},{\boldsymbol{h}}^{f'}_{\text{dec}, t}]$ which incorporates shared and domain-specific features.

\subsection{Dynamic Fusion for Querying Knowledge} \label{sec:moe}
The shared-private framework can capture the corresponding specific feature, but neglects the fine-grained relevance across certain subsets of domains.
We further propose a dynamic fusion layer to explicitly leverage all domain knowledge, which is shown in Figure~\ref{fig:knowledge_transfer}.
Given an instance from any domain, we first put it to multiple private encoder-decoder to obtain domain-specific features from all domains.
Next, all domain-specific features are fused by a dynamic domain-specific feature fusion module, followed by a shared-specific feature fusion for obtaining shared-specific features.
\paragraph{Dynamic Domain-Specific Feature Fusion}
Given domain-specific features from all domains, a Mixture-of-Experts mechanism (MoE) \cite{guo-etal-2018-multi} is adapted to dynamically incorporate all domain-specific knowledge for the current input in both encoder and decoder. 
Now, we give a detailed description on how to fuse the timestep $t$ of decoding and the fusion process is the same to encoder.
Given all domain feature representations in $t$ decoding steps: $\{\boldsymbol{h}_{{\text{dec}, t}}^{d_{i}}\}_{i=1}^{|\mathbb {D}|}$, where $|\mathbb {D}|$ represents the number of domains,
an \emph{expert gate} $E$ takes $\{\boldsymbol{h}_{{\text{dec}, t}}^{d_{i}}\}$ as input and outputs a softmax score $\alpha_{t, i}$ that represents the degree of correlation between each domain and the current input token.
We achieve this by a simple feedforward layer:
\begin{eqnarray} \label{moe}
	\boldsymbol{\alpha}_t = \text{Softmax}( \boldsymbol{W}*\boldsymbol{h}_{{\text{dec}, t}}^{d} + \boldsymbol{b}).
\end{eqnarray}

The final domain-specific feature vector is a mixture of all domain outputs, dictated by the expert gate weights ${\boldsymbol{\alpha}_t} = (\alpha_{t, 1},\dots,\alpha_{t, \mathbb{|D|}})$, which can be written as $\boldsymbol{h}_{\text{dec}, t}^{{d_{f}}} = \sum_i \alpha_{t, i} \boldsymbol{h}_{\text{dec}, t}^{d_{i}}$.

During training, take the decoder for example, we apply the cross-entropy loss $L_\text{dec}^{moe}$ as the supervision signal for the expert gate to predict the domain of each token in the response, where the expert gate output $\alpha_{t}$ can be treated as the $t^{th}$ token's predicted domain probability distribution by multiple private decoder.
Hence, the more accurate the domain prediction is, the more correct expert gets:
\begin{eqnarray}
\begin{aligned}
L_\text{dec}^{moe} &= -\sum_{t=1}^{n}\sum_{i=1}^{\mathbb{|D|}} (e_{i} \cdot \log (\alpha_{t,i} | \boldsymbol{\theta_{s}},\boldsymbol{\theta_{dec}^m})),
\end{aligned}
\end{eqnarray}
where $\boldsymbol{\theta_{s}}$ represents the parameters of encoder-decoder model, $\boldsymbol{\theta_{dec}^{m}}$ represents the parameters of the MoE module (Eq.~\ref{moe}) in the decoder and $e_i \in \{0,1\}$ represents whether the response with $n$ tokens belongs to the domain $d_i$.
Similarly, we can get the $L_\text{enc}^{moe}$ for the encoder and sum up them as: $\mathcal{L}_{moe} = L_\text{enc}^{moe} + L_\text{dec}^{moe}$.

$\mathcal{L}_{moe}$ is used to encourage samples from a certain source domain to use the correct expert, and each expert learns corresponding domain-specific features. 
When a new domain has little or no labeled data, the expert gate can automatically calculate the correlation between different domains with the target domain and thus better transfer knowledge from different source domains in both encoder and decoder module.

\paragraph{Shared-Specific Feature Fusion}
We directly apply $\shprivate$ operation to fuse shared and final domain-specific feature:
\begin{eqnarray}
\shprivate : (\boldsymbol{h}_{{\text{dec}, t}}^{s}, \boldsymbol{h}_{{\text{dec}, t}}^{d_{f}}) \rightarrow \boldsymbol{h}_{{\text{dec}, t}}^{f}.
\end{eqnarray} 

Finally, we denote the dynamic fusion function as $\dynfusion ({\boldsymbol{h}_{\text{dec,t}}^{s}}, \{{\boldsymbol{h}}_\text{dec,t}^{d_{i}}\}_{i=1}^{|\mathbb{D}|})$. Similar to Section~\ref{sec:shared-private}, we replace $[\boldsymbol{h}^{}_{\text{dec},t}, \boldsymbol{h}^{'}_{\text{dec}, t}]$ in Eq.~\ref{decoder_initial} with $[ \boldsymbol{h}_{{\text{dec}, t}}^{f},{\boldsymbol{h}}^{f'}_{\text{dec}, t}]$. The other components are kept the same as the \textit{shared-private encoder-decoder} framework.

\paragraph{Adversarial Training}
To encourage the model to learn domain-shared features, we apply adversarial learning on the shared encoder and decoder module.
Following \newcite{liu-etal-2017-adversarial}, a gradient reversal layer \cite{ganin2014unsupervised} is introduced after the domain classifier layer.
The adversarial training loss is denoted as $\mathcal{L}_{adv}$.
 We follow \newcite{qin-etal-2019-stack} and the final loss function of our Dynamic fusion network is defined as:
\begin{equation}
\mathcal{L} = \gamma_{b}\mathcal{L}_{basic} + \gamma_{m}\mathcal{L}_{moe} + \gamma_{a}\mathcal{L}_{adv},
\end{equation}
where $\mathcal{L}_{basic}$ keep the same as GLMP \cite{DBLP:conf/iclr/WuSX19}, $\gamma_{b}$, $\gamma_{m}$ and $\gamma_{a}$ are hyper-parameters.
 More details about $\mathcal{L}_{basic}$ and $\mathcal{L}_{adv}$ can be found in appendix.
\begin{table}[t]
	\begin{center}
		
		\scalebox{0.65}{
			\begin{tabular}{l|l|l|l|l}
				\hline 
				\textbf{Dataset} &\textbf{ Domains}&\textbf{Train} & \textbf{Dev} & \textbf{Test} \\ 
				\hline
				\multirow{1}*{SMD} 
				&Navigate, Weather, Schedule& 2,425 & 302 &304 \\	
				\hline
				\multirow{1}*{Multi-WOZ 2.1} 
				&Restaurant, Attraction, Hotel & 1,839 & 117 & 141 \\
				\hline
			\end{tabular}
		}
	\end{center}
	\caption{Statistics of datasets.}
	\label{tab:OverallSta}
\end{table}

\begin{table*}[ht]
	\centering
	\begin{adjustbox}{width=\textwidth}
		\begin{tabular}{l||ccccc||ccccc}
			\hline
			& \multicolumn{5}{c}{SMD}  & \multicolumn{5}{c}{Multi-WOZ 2.1} \\
			\hline \hline
			Model & BLEU & F1 & 
			\begin{tabular}[c]{@{}l@{}}Navigate\\~ ~~ F1\end{tabular} & \begin{tabular}[c]{@{}c@{}}Weather \\F1\end{tabular} & \begin{tabular}[c]{@{}c@{}}Calendar \\F1\end{tabular} &
			BLEU & F1 & 
			\begin{tabular}[c]{@{}c@{}}Restaurant \\F1\end{tabular} & 
			\begin{tabular}[c]{@{}c@{}}Attraction \\F1\end{tabular} & 
			\begin{tabular}[c]{@{}c@{}}Hotel \\F1\end{tabular} 
			\\ 
			\hline
			{Mem2Seq} \cite{madotto-etal-2018-mem2seq} & 12.6 & 33.4 & 20.0 & 32.8 & 49.3  & 6.6 & 21.62 & 22.4 & 22.0 & 21.0  \\ 
			{DSR} \cite{wen-etal-2018-sequence} & 12.7 & 51.9 & 52.0 & 50.4 & 52.1 & {9.1} & 30.0 & 33.4 & 28.0 & 27.1  \\ 
			{KB-retriever} \cite{qin-etal-2019-entity} & 13.9 & 53.7 & 54.5 & 52.2 & 55.6 & - & - & - & - & -  \\ 
			{GLMP}  \cite{DBLP:conf/iclr/WuSX19} & 13.9 & 60.7 & 54.6 & 56.5 & 72.5 & 6.9 & 32.4 & 38.4 & 24.4 & 28.1  \\
			\hline
			{{Shared-Private framework} (Ours)} & {13.6} & {61.7} & {56.3} & {56.5} & {72.8} & {6.6} & {33.8} & {39.8} & {26.0} & {28.3}  \\ 
			\hdashline
			{Dynamic Fusion framework (Ours)} & \textbf{14.4*} & \textbf{62.7*} & \textbf{57.9*} & \textbf{57.6*} & \textbf{73.1*} & \textbf{9.4*} & \textbf{35.1*} & \textbf{40.9*} & \textbf{28.1*} & \textbf{30.6*}  \\ \hline 
		\end{tabular}
	\end{adjustbox}
	\caption{Main results. The numbers with * indicate that the improvement of our framework over all baselines is statistically significant with $p<0.05$ under t-test.} \label{tab:results}
	\vspace{-0.3cm}
\end{table*}

\section{Experiments}
\subsection{Datasets}
Two publicly available datasets are used in this paper, which include SMD \cite{eric-etal-2017-key}
 and an extension of Multi-WOZ 2.1 \cite{budzianowski-etal-2018-multiwoz} that we equip the corresponding KB to every dialogue.\footnote{The constructed datasets will be publicly available for further research.}
The detailed statistics are also presented in  Table~\ref{tab:OverallSta}.
 We follow the same partition as \newcite{eric-etal-2017-key}, \newcite{madotto-etal-2018-mem2seq} and \newcite{DBLP:conf/iclr/WuSX19} on SMD and \cite{budzianowski-etal-2018-multiwoz} on Multi-WOZ 2.1.

\subsection{Experimental Settings}
The dimensionality of the embedding and LSTM hidden units is $128$. The dropout ratio we use in our framework is selected from $\{0.1,0.2\}$ and the batch size from  $\{16,32\}$. In the framework, we adopt the weight typing trick \cite{DBLP:conf/iclr/WuSX19}.
We use Adam \cite{DBLP:journals/corr/KingmaB14} to optimize the parameters in our model and adopt the suggested hyper-parameters for optimization.
All hyper-parameters are selected according to validation set.
More details about hyper-parameters can be found in Appendix.

\subsection{Baselines}
We compare our model with the following state-of-the-art baselines.
\begin{itemize}
	\item  \textbf{Mem2Seq} \cite{madotto-etal-2018-mem2seq}:
	the model takes dialogue history and KB entities
	as input and uses a pointer gate
	to control either generating a vocabulary word or selecting 
	an input as the output.
	\item \textbf{DSR} \cite{wen-etal-2018-sequence}: the model leverages dialogue state representation to retrieve the KB implicitly and  applies copying mechanism to retrieve entities from knowledge base while decoding.
	\item  \textbf{KB-retriever} \cite{qin-etal-2019-entity}: the model adopts a retriever module to retrieve the most relevant KB row and filter the irrelevant information for the generation process.
	\item  \textbf{GLMP} \cite{DBLP:conf/iclr/WuSX19}:  the framework adopts the global-to-local pointer mechanism to query the knowledge base during decoding and achieve state-of-the-art performance.
\end{itemize}

For \textit{Mem2Seq}, \textit{DSR}, \textit{KB-retriever} \footnote{For Multi-WOZ 2.1 dataset, most dialogs are supported by more than single row, which can not processed by \textit{KB-retriever}, so we compare our framework with it in SMD and Camrest datasets.}, we adopt the reported results from \newcite{qin-etal-2019-entity} and \newcite{DBLP:conf/iclr/WuSX19}.
For \textit{GLMP}, we rerun their public code to obtain results on same datasets.\footnote{Note that, we find that \newcite{DBLP:conf/iclr/WuSX19} report macro entity F1 as the micro F1, so we rerun their models (https://github.com/jasonwu0731/GLMP) and obtain results.} \

\begin{table}[t]
	\centering
	\begin{adjustbox}{width=0.45\textwidth}
		\begin{tabular}{l|c|c}
			\hline
			\multirow{2}{*}{\textbf{Model}} & \multicolumn{2}{c}{\textbf{Entity F1 (\%)}} \\ \cline{2-3} 
			& \textbf{Test} & $\boldsymbol\Delta$ \\ \hline \hline
			Full model & \textbf{62.7} & - \\ \hline
			\hline
			~~w/o Domain-Shared Knowledge Transfer & 59.0 & 3.7 \\ \hline
			~~w/o Dynamic Fusion Mechanism & 60.9 & 1.8 \\ \hline
			~~w/o Multi-Encoder & 61.0 & 1.7 \\ \hline
			~~w/o Multi-Decoder & 58.9 & 3.8 \\ \hline
			~~w/o Adversarial Training & 61.6 & 1.1 \\ \hline
		
		\end{tabular}
	\end{adjustbox}
	\caption{Ablation tests on the
		SMD test set. 
	}
	\label{tab:ablation}
\end{table}

\subsection{Results}
Follow the prior work \cite{eric-etal-2017-key, madotto-etal-2018-mem2seq,wen-etal-2018-sequence, DBLP:conf/iclr/WuSX19,qin-etal-2019-entity}, we adopt the \textit{BLEU} and \textit{Micro Entity F1} metrics to evaluate model performance.
The results on the two datasets are shown in Table~\ref{tab:results}, we can observe that:
1) The basic \textit{shared-private} framework outperforms the best prior model GLMP in all the datasets.
This indicates that the combination of domain-shared and domain-specific features can better enhance each domain performanc compared with only utilizing the implicit domain-shared features.
2) Our framework achieves the state-of-the-art performance on two multi-domain task-oriented dialog datasets, namely SMD and Multi-WOZ 2.1.
On SMD dataset, our model has the highest BLEU compared with baselines, which shows that our framework can generate more fluent response.
More importantly, our model outperforms GLMP by 2.0\% overall, 3.3\% in the Navigate domain, 1.1\% in the Weather domain and 0.6\% in Schedule domain on entity F1 metric, which indicates that considering relevance between target domain input and all domains is effective for enhancing performance of each domain.
On Multi-Woz 2.1 dataset, the same trend of improvement has been witnessed, which further shows the effectiveness of our framework.

\subsection{Analysis}
We study the strengths of our model from several perspectives on SMD dataset. We first conduct several ablation experiments to analyze the effect of different components in our framework. Next, we conduct domain adaption experiments to verify the transferability of our framework given a new domain with little or no labeled data. 
In addition, we provide a visualization of the dynamic fusion layer and case study to better understand how the module affects and contributes to the performance. 
\begin{figure*}[htbp]
	\centering
	\begin{adjustbox}{width=0.9\textwidth}
		\begin{subfigure}[t]{0.3\linewidth}
			\includegraphics[width=\linewidth]{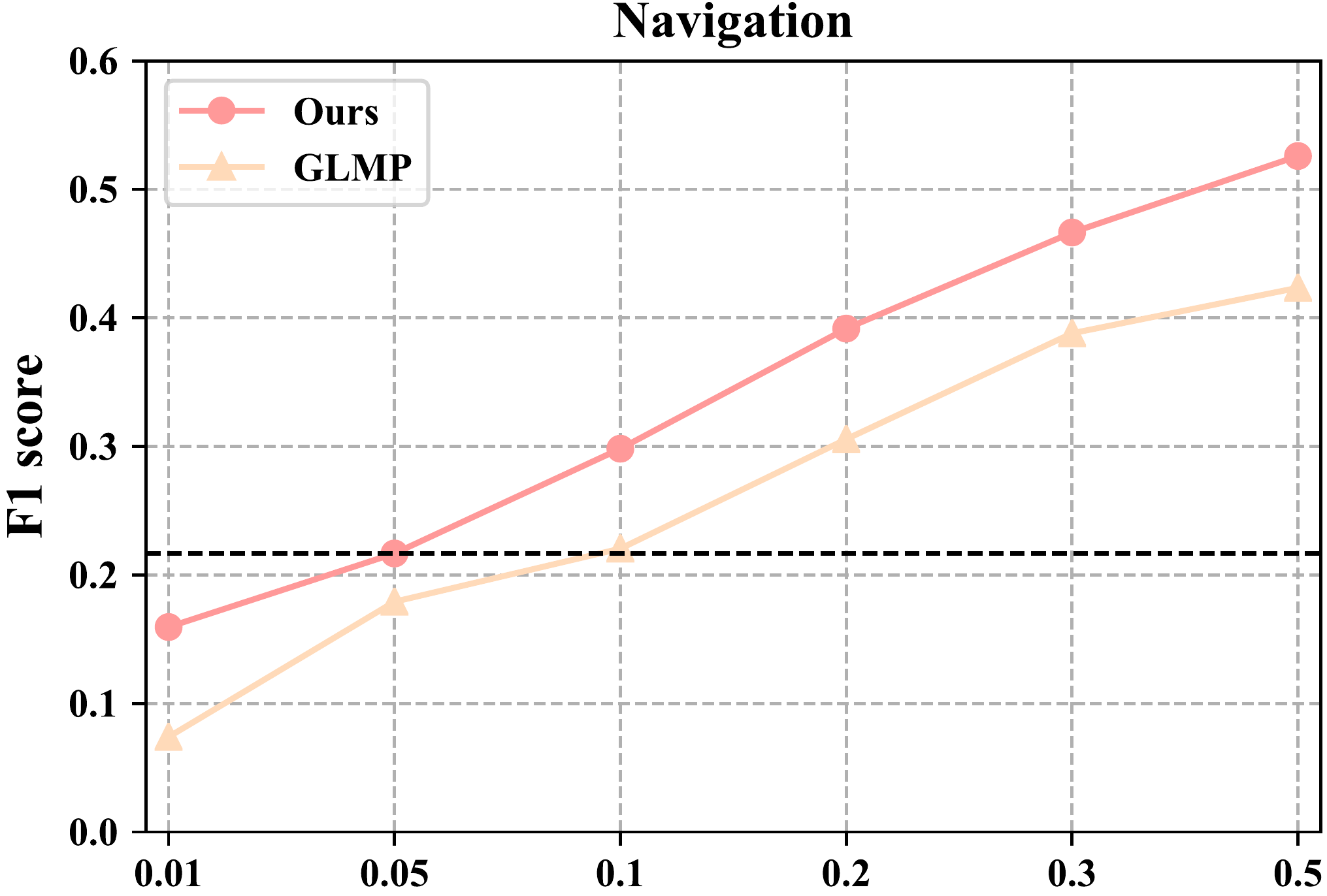}
			\caption{Navigate Domain}
			\label{fig:myfig3}
		\end{subfigure}
		\quad
		\begin{subfigure}[t]{0.3\linewidth}
			\includegraphics[width=\linewidth]{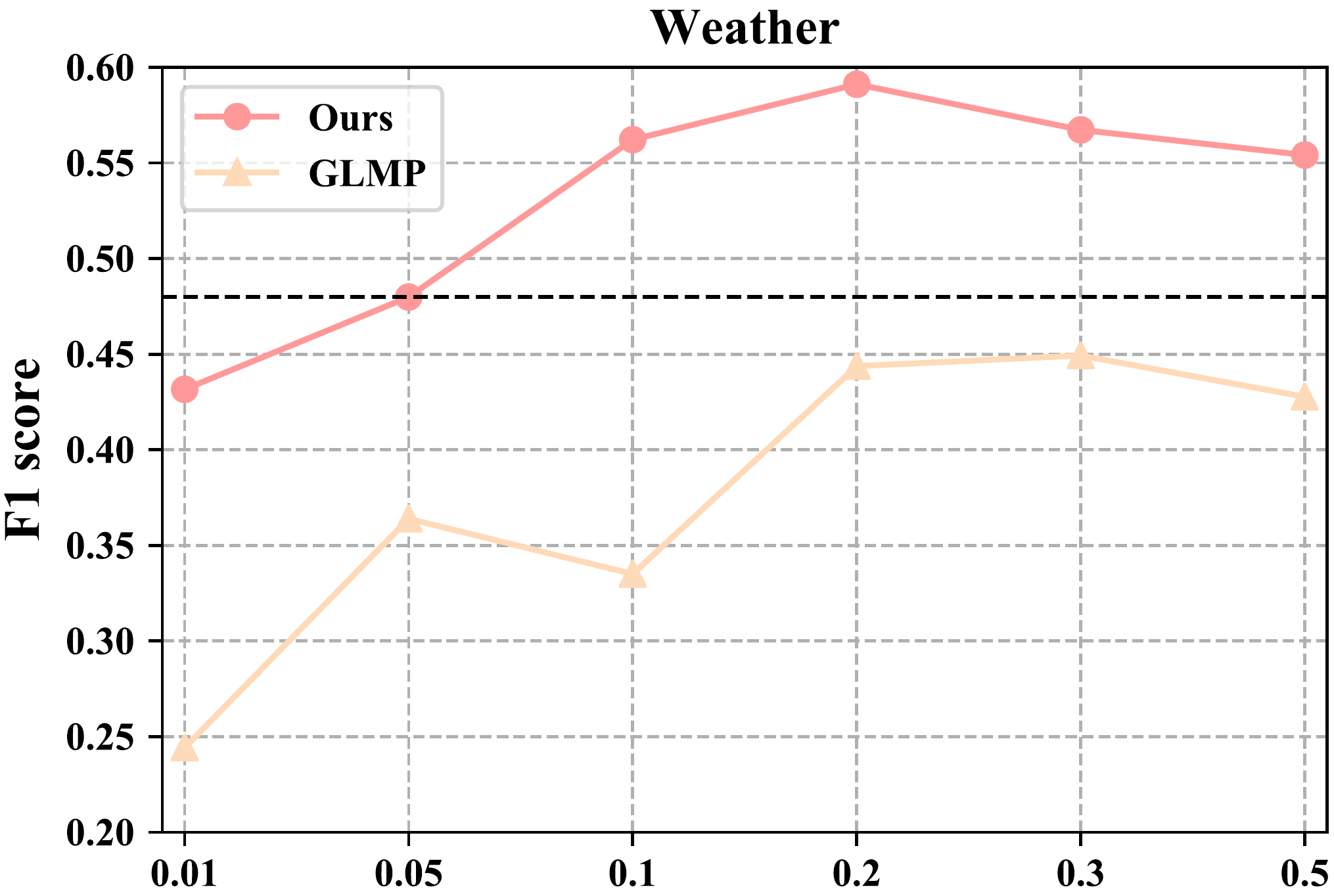}
			\caption{Weather Domain}
			\label{fig:myfig4}
		\end{subfigure}
		\quad
		\begin{subfigure}[t]{0.3\linewidth}
			\includegraphics[width=\linewidth]{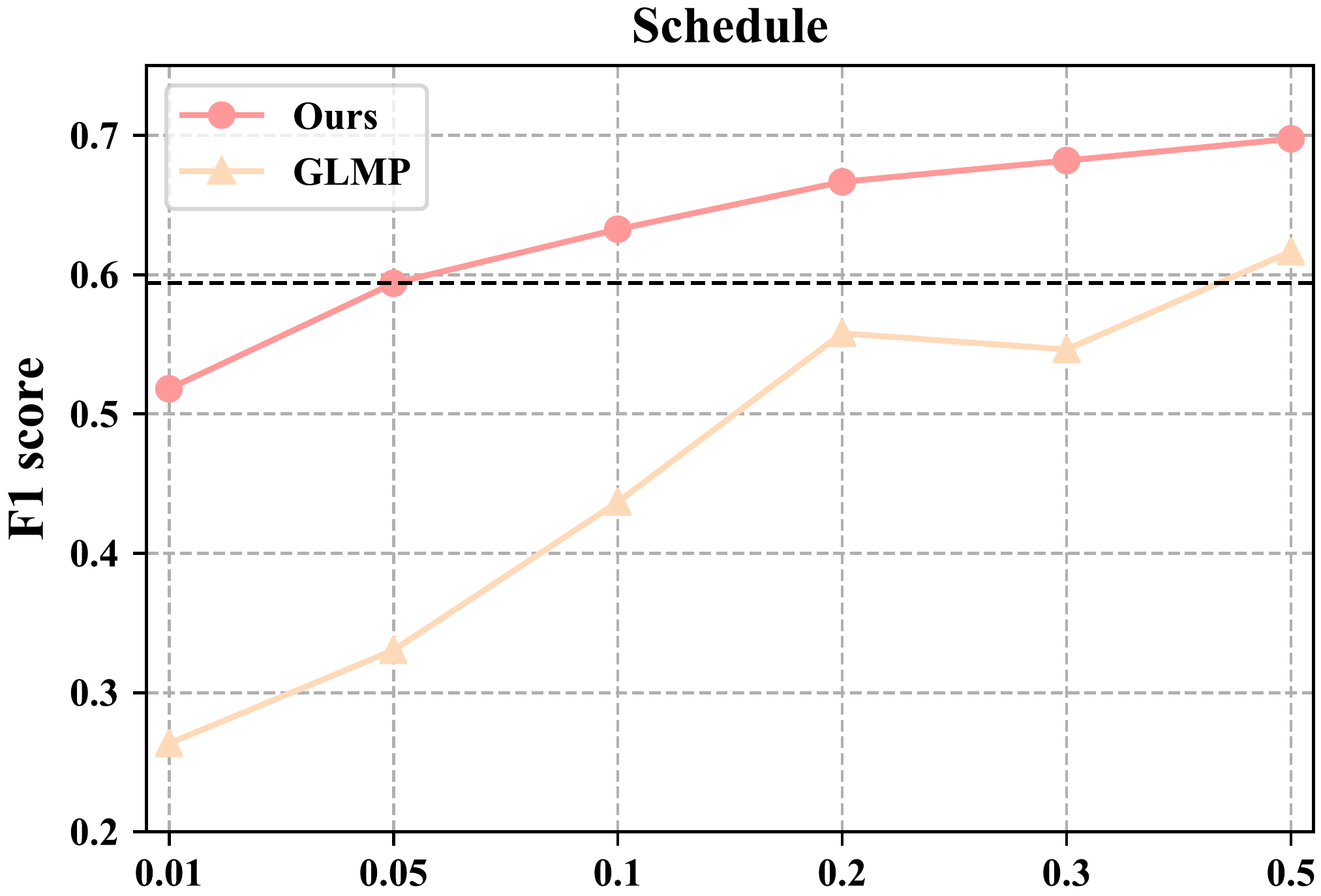}
			\caption{Schedule Domain}
			\label{fig:myfig5}
		\end{subfigure}
	\end{adjustbox}
	\caption{Performance of domain adaption on different subsets of original training data.}
	\label{fig:few-shot}
\end{figure*}

\begin{figure}[t]
	\centering
	\includegraphics[scale=0.25]{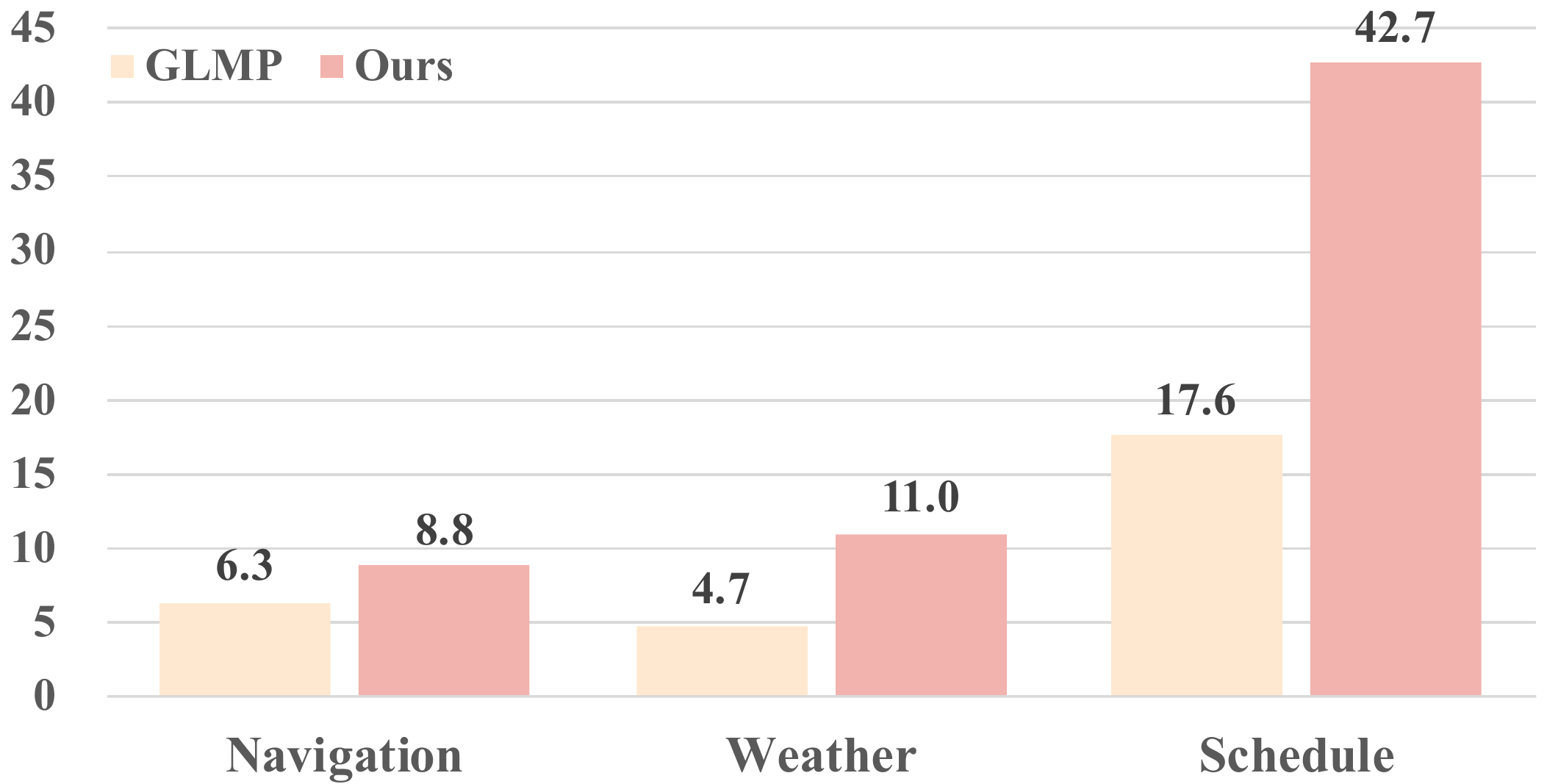}
	\caption{
		Zero-shot performance (F1 score) on each domain on SMD dataset.
		The x-axis domain name represents that the domain is unseen and other two domains is the same as original dataset.
	}
	\label{fig:zero-shot}
\end{figure}

\begin{table}[t]
	\centering
	\begin{adjustbox}{width=0.4\textwidth}
		\begin{tabular}{rccc}
			\hline
			Model& Correct& Fluent & Humanlike\\\hline
			GLMP & 3.4 & 3.9 & 4.0 \\
	
			Our framework & 3.6 & 4.2 & 4.2\\
			\hline
			Agreement & 53\% & 61\% & 74\%\\
			\hline
		\end{tabular}
	\end{adjustbox}
	\caption{
		Human evaluation of responses on the randomly selected dialogue history.
	}
	\label{tab:human}
\end{table}

\subsubsection{Ablation}
Several ablation experiments and results are shown in Table~\ref{tab:ablation}.
In detail, 1) w/o Domain-shared Knowledge Transfer denotes that we remove domain-shared feature and just keep fused domain-specific feature for generation.
2) w/o Domain Fusion mechanism denotes that we simply sum all domain-specific features rather than use the MOE mechanism to dynamically fuse domain-specific features.
3) w/o Multi-Encoder represents that we remove multi-encoder module and adopt one shared encoder in our framework.
4) w/o Multi-Decoder represents that we remove the multi-decoder module and adopt one shared decoder.
5) w/o Adversarial Training denotes that we remove the adversarial training in experimental setting.
Generally, all the proposed components are effective to contribute the final performance.
Specifically, we can clearly observe the effectiveness of our dynamic fusion mechanism where w/o domain-specific knowledge fusion causes 1.8\% drops and the same trend in removing domain-shared knowledge fusion.
This further verifies that domain-shared and domain-specific feature are benefit for each domain performance.
\begin{figure}[t]
	\centering
	\includegraphics[scale=0.25]{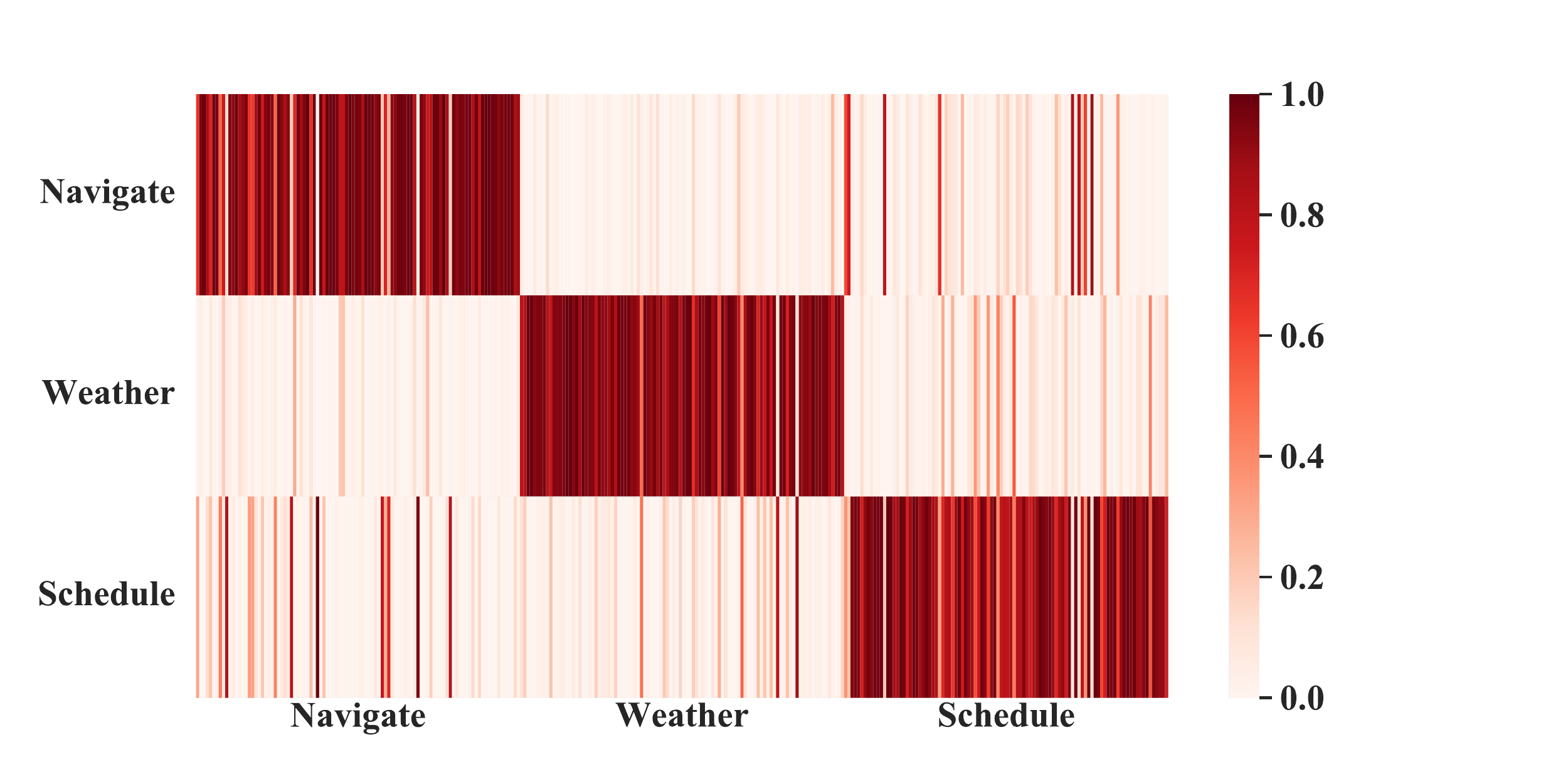}
	\caption{
		Distribution of Mix-of-the-expert mechanism across source domains for randomly selected 100 examples in each domain on SMD dataset.
	}
	\label{fig:visualize}
\end{figure}
\subsubsection{Domain Adaption}
\paragraph{Low-Resource Setting}
To simulate low-resource setting, we keep two domains unchanged, and the ratio of the except domain from original data varies from [1\%, 5\%, 10\%, 20\%, 30\%, 50\%].
The results are shown in Figure~\ref{fig:few-shot}.
We can find that:
(1) Our framework outperforms the GLMP baseline on all ratios of the original dataset.
When the data is only 5\% of original dataset, our framework outperforms GLMP by 13.9\% on all domains averagely.
(2) Our framework trained with 5\% training dataset can achieve comparable and even better performance compared to GLMP with 50\% training dataset on some domains.
This implies that our framework effectively transfers knowledge from  other domains to achieve better performance for the low-resources new domain.
\paragraph{Zero-Shot Setting}
Specially, we further evaluate the performance of domain adaption ability on the zero-shot setting given an unseen domain. 
We randomly remove one domain from the training set, and other domain data remained unchanged to train the model.
During test, the unseen domain input use the MoE to automatically calculate the correlation between other domains and the current input and get the results. 
Results are shown in Figure \ref{fig:zero-shot}, we can see our model significantly outperforms GLMP on three domains, which further demonstrate the transferability of our framework.

\subsubsection{Visualization of Dynamic Fusion Layer}
To better understand what our dynamic fusion layer has learned, we visualize the gate distribution for each domain in low-resource (5\%) setting, which fuses domain-specific knowledge among various cases. 
As shown in the Figure~\ref{fig:visualize}, for a specific target domain, different examples may have different gate distributions, which indicates that our framework successfully learns how to transfer knowledge between different domains. 
For example, the navigation column contains 100 examples from its test set and each row show the corresponding expert value. 
More specifically, in the navigation column, we observe that the expert value in schedule domain is bigger than weather domain, which indicates schedule domain transfers more knowledge to navigation than weather domain.
\begin{figure}[t]
	\centering
	\includegraphics[scale=0.38]{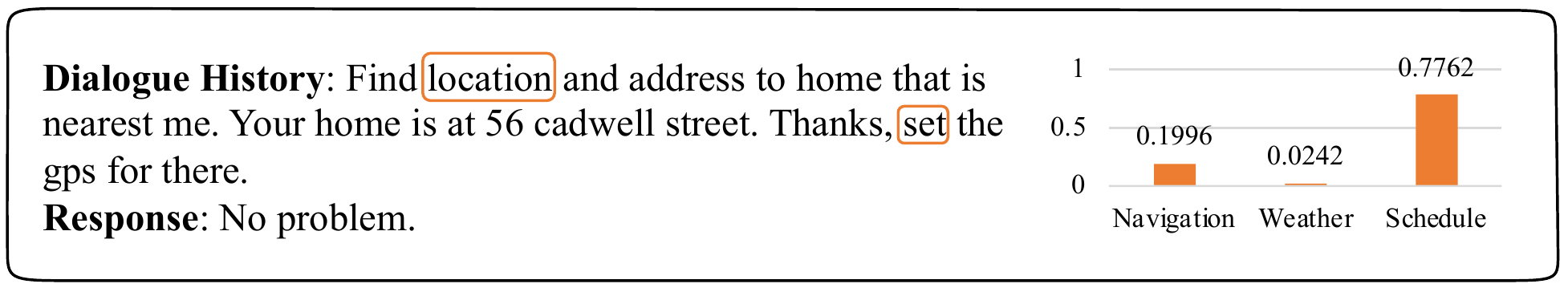}
	\caption{
		Case of of expert gate distribution in SMD dataset. Text segments with red color represents appearing in both schedule and navigation domain.
	}
	\label{fig:visualize-case}
\end{figure}

\subsubsection{Case Study}
Furthermore, we provide one case for navigation domain and their corresponding expert gate distribution.
The cases are generated with 5\% training data in the navigation domain and other two domain datasets keep the same, which can better show how the other two domains transfer knowledge to the low-resource domain. 
As shown in Figure~\ref{fig:visualize-case},  the expert value of schedule domain is bigger than the weather domain, which indicates the schedule contributes more than weather domain.
In further exploration, we find word ``\textit{location}''  and ``\textit{set}'' appear both in navigation and schedule domain, which shows schedule has closer relation with navigation than weather, which indicates our model successfully transfers knowledge from the closest domain.

\subsubsection{Human Evaluation}
We provide human evaluation on our framework and other baseline models.
We randomly generated 100 responses. These responses are based on distinct dialogue history on the SMD test data.
Following \newcite{wen-etal-2018-sequence} and \newcite{qin-etal-2019-entity}, We hired human experts and asked them to judge the quality of the responses according to correctness, fluency, and humanlikeness on a scale from 1 to 5.

Results are illustrated in Table~\ref{tab:human}.
 We can see that our framework outperforms GLMP on all metrics, which is consistent with the automatic evaluation.

\section{Related Work}
Existing end-to-end task-oriented systems can be classified into two main classes.
A series of work trains a single model on the mixed multi-domain dataset. 
\newcite{eric-etal-2017-key} augments the vocabulary distribution by concatenating KB attention to generatge entities.
\newcite{lei-etal-2018-sequicity} first integrates track dialogue believes in end-to-end task-oriented dialog.
\newcite{madotto-etal-2018-mem2seq} combines end-to-end memory network \cite{DBLP:conf/nips/SukhbaatarSWF15} into sequence generation.
\newcite{gangi-reddy-etal-2019-multi} proposes a multi-level memory architecture which first addresses queries, followed by results and finally each key-value pair within a result.
\newcite{DBLP:conf/iclr/WuSX19} proposes a global-to-locally pointer mechanism to query the knowledge base.
Compared with their models, our framework can not only explicitly utilize domain-specific knowledge but also consider different relevance between each domain.
Another series of work trains a model on each domain separately.
\newcite{wen-etal-2018-sequence} leverages dialogue state representation to retrieve the KB implicitly.
\newcite{qin-etal-2019-entity} first adopts the KB-retriever to explicitly query the knowledge base. 
Their works consider only domain-specific features. In contrast, our framework explicitly leverages domain-shared features across domains.

The shared-private framework has been explored in many other task-oriented dialog components.
\newcite{liu2017multi} applies a shared-private LSTM to generate shared and domain-specific features.
\newcite{zhong-etal-2018-global} proposes a global-local architecture to learn shared feature across all slots and specific feature for each slot.
More recently,
\newcite{zhang-etal-2018-shaped}
utilizes the shared-private model for text style adaption.
In our work, we explore shared-private framework in end-to-end task-oriented dialog to better transfer domain knowledge for querying knowledge base.
In addition, we take inspiration from \newcite{guo-etal-2018-multi}, who successfully apply the mix-of-the-experts (MoE) mechanism in multi-sources domain and cross-lingual adaption tasks.
Our model not only combines the strengths of MoE to incorporate domain-specific feature, but also 
applies adversarial training to encourage generating shared feature.
To our best of knowledge, we are the first to effectively explore shared-private framework in multi-domain end-to-end task oriented dialog.
\section{Conclusion}
In this paper, we propose to use a shared-private model to investigate explicit modeling domain knowledge for multi-domain dialog.
In addition, a dynamic fusion layer is proposed to dynamically capture the correlation between a target domain and all source domains.
Experiments on two datasets show the effectiveness of the proposed models.
Besides, our model can quickly adapt to a new domain with little annotated data.

\section*{Acknowledgements}
We thank Min Xu, Jiapeng Li, Jieru Lin and Zhouyang Li for their insightful discussions.
We
also thank all anonymous reviewers for their constructive comments. 
This work was supported by the National Natural Science Foundation of China (NSFC) via grant 61976072, 61632011 and 61772153. Besides, this work also faxed the support via Westlake-BrightDreams Robotics research grant.

\bibliography{acl2020}
\bibliographystyle{acl_natbib}

\appendix
\clearpage

\section{Appendices}
\label{sec:appendix}

\subsection{Hyperparameters Setting}

The hyperparameters used for SMD and Multi-WOZ 2.1 dataset are shown in Table \ref{fig:hyper}.

\begin{table}[t]
	\scalebox{0.74}{
		\begin{tabular}{l|c|c}
			\hline
			\textbf{Hyperparameter Name} & \textbf{SMD} & \textbf{Multi-WOZ 2.1} \\
			\hline
			Batch Size                     & 16    & 32            \\
			Hidden Size                    & 128   & 128           \\
			Embedding Size                 & 128   & 128           \\
			Learning Rate                  & 0.001 & 0.001         \\
			Dropout Ratio                  & 0.2   & 0.1           \\
			Teacher Forcing Ratio          & 0.9   & 0.9           \\
			Number of Memory Network's Hop & 3     & 3             \\
			\hline
		\end{tabular}
	}
	\caption{Hyperparameters we use for SMD and Multi-WOZ 2.1 dataset.}\label{fig:hyper}
\end{table}

\subsection{Basic Loss Function}

The loss $\mathcal{L}_{basic}$ used in our \textit{Shared-Private Encoder-Decoder Model} is the same as GLMP.
Different with the standard sequence-to-sequence with attention mechanism model, we use $[\boldsymbol{h}_{\text{dec}, t}^{f}, \boldsymbol{h}_{\text{dec}, t}^{f^{'}}]$ to replace $[\boldsymbol{h}_{\text{dec}, t}^{}, \boldsymbol{h}_{\text{dec}, t}^{'}]$ and then get the sketch word probability distribution $P_t^{vocab}$. Based on the gold sketch response $\boldsymbol{Y}^s = (y_1^s, \dots, y_n^s)$, we calculate the standard cross-entropy loss $\mathcal{L}_{v}$ as follows: 
\begin{align}
P_t^{vocab} &= \operatorname{Softmax}(\boldsymbol{U}[\boldsymbol{h}^{f}_{\text{dec}, t}, \boldsymbol{h}^{f^{'}}_{\text{dec}, t}]), \\
\mathcal{L}_{v}&=\sum_{t=1}^{n}-log(P_t^{vocab}(y_t^s)).
\end{align}

Given the system response $Y$, we get the global memory pointer label sequence $G^{label} = (\hat g_1, \dots, \hat g_{b+T})$ and local memory pointer label sequence $L^{label} = (\hat l_1, \dots, \hat l_n)$ as follows: 
\begin{align}
\hat g_{i}&\!=\!\left\{\begin{array}{ll}{1} & {\text { if } \operatorname{Object}(m_{i}) \in Y} \\ {0} & {\text { otherwise }}\end{array}\right., \\
\hat{l}_{t}&\!=\!\left\{\begin{array}{ll}{\max (z)} & {\text { if } \exists z \text { s.t. } y_{t}\!=\!\operatorname{Object}(m_{z})} \\ {b+T+1} & {\text { otherwise }}\end{array}\right.,
\end{align}
where $m_i$ represents one triplet in the external knowledge $M=[B;X]=(m_1, \dots, m_{b+T})$ and $\operatorname{Object}(\cdot)$ function is denoted as getting the object word from a triplet.

Then, the $\mathcal{L}_{g}$ can be written as follows:
\begin{equation}
\mathcal{L}_{g}\!=\!-\sum_{i=1}^{b+T}\left(\hat g_{i} \cdot \log g_{i} + (1-\hat g_{i}) \cdot \log \left(1-g_{i}\right)\right). 
\end{equation}

Based on the $L^{label}$ and $P_t=(p_1^k, \dots, p_{b+T}^k)$, we can calculate the standard cross-entropy loss $\mathcal{L}_{l}$ as follows:
\begin{equation}
\mathcal{L}_{l}=\sum_{t=1}^{n}-\log (P_t(\hat l_{t})).
\end{equation}

Finally, $\mathcal{L}_{basic}$ is the weighted-sum of three losses:
\begin{equation}
\mathcal{L}_{basic} = \gamma_{g} \mathcal{L}_{g} + \gamma_{v} \mathcal{L}_{v} + \gamma_{l} \mathcal{L}_{l},
\end{equation}
where $\gamma_{g}$, $\gamma_{v}$ and $\gamma_{l}$ are hyperparameters.

\subsection{Adversarial Training}
We apply a Convolutional Neural Network (CNN) as domain classifier both in the shared encoder and shared decoder to identify the domain of shared representation of dialogue history $\boldsymbol{H}_{\text{enc}}^{s}$ and response $\boldsymbol{H}_{\text{dec}}^{s}$. Take the encoder for example, based on the $\boldsymbol{H}_{\text{enc}}^{s}$, we can extract the context representation $\boldsymbol{c}_{\text{enc}}^s$ by CNN and then $\boldsymbol{\beta}_{\text{enc}} \in \mathbb{R^{|D|}}$ can be calculated as follows:
\begin{equation}
	\boldsymbol{\beta}^{}_{\text{enc}} \!=\! \operatorname{Sigmoid}(\operatorname{LeakyReLU}(\boldsymbol{W}^{}_{\text{enc}}(\boldsymbol{c}^s_{\text{enc}})),
\end{equation}

Then we train the domain classifier by optimizing its parameters $\boldsymbol{\theta_{d}}$ to minimize the sequence-level binary cross-entropy loss $L_\text{enc}^{adv}$ as follows:
\begin{equation}
\begin{aligned}
\mathop{\max}_{\boldsymbol{\theta_{s}}} \mathop{\min}_{\boldsymbol{\theta_{d}}} L_\text{enc}^{adv} &\!=\!  -\sum_{i=1}^{\mathbb{|D|}} (e_{i} \cdot \log (\beta_{\text{enc}, i} | \boldsymbol{\theta_{s}},\boldsymbol{\theta_{d}}) \\ &\!+\! (1 - e_{i}) \cdot \log (1 - \beta_{\text{enc}, i}|\boldsymbol{\theta_{s}},\boldsymbol{\theta_{d}})), 
\end{aligned}
\end{equation}
where $\beta_{\text{enc}, i}$ represents the probability of the input dialogue history belongs to the domain $d_i$. Similarly, we can get the $L_\text{dec}^{adv}$ and sum up them as: $\mathcal{L}_{adv} = L_\text{enc}^{adv} + L_\text{dec}^{adv}$.

In order to update the encoder-decoder model parameters $\boldsymbol{\theta_{s}}$ underlying the domain classifier, we introduce the gradient reversal layer to reverse the gradient direction which trains our model to extract domain-shared features to confuse the classifier. On the one hand, we train the domain classifier to minimize the domain classification loss. On the other hand, we update the parameters of the network underlying the domain classifier to maximize the domain classification loss, which works adversarially towards the domain classifier. This encourages that our shared encoder and decoder are trained to extract domain-shared features. 
\end{document}